%% file: CED.tex
\documentclass[pdflatex,sn-mathphys-num]{sn-jnl}

\usepackage{amssymb}

\usepackage{lineno}

\usepackage{url}
\let\orcidlogo\relax
\usepackage{orcidlink}

\usepackage[acronym,toc]{glossaries}
\usepackage{colortbl,hhline}
\usepackage{float}
\usepackage{comment}
\usepackage{gensymb}
\usepackage{multirow}
\usepackage{glossaries}
\usepackage{graphicx}
\usepackage{adjustbox}
 \usepackage{booktabs}
\loadglsentries{glossary}
\makeglossaries
\usepackage{soul}
\usepackage{lineno}
\usepackage{xcolor}
\hypersetup{
 colorlinks,
 linkcolor={red!50!black},
 citecolor={blue!50!black},
 urlcolor={blue!80!black}
}

\makeglossaries

\theoremstyle{thmstyleone}%
\theoremstyle{thmstyletwo}%

\theoremstyle{thmstylethree}%

\begin{document}
\title[\acrfull*{ced} for Real-Time Environmental Monitoring in Aquatic Ecosystems]{\acrfull*{ced} for Real-Time Environmental Monitoring in Aquatic Ecosystems}

\author[1]{\fnm{Dennis \orcidlink{0009-0000-7670-3659}} \sur{Monari}\orcidlink{0009-0000-7670-3659}}\email{dennis.monari@ntu.ac.uk}

\author[1]{\fnm{Farhad} \sur{Fassihi Tash}\orcidlink{0000-0002-6200-5179}}\email{farhad.fassihitash@ntu.ac.uk}

\author[1]{\fnm{Jordan J.} \sur{Bird}\orcidlink{0000-0002-9858-1231}}\email{jordan.bird@ntu.ac.uk}

\author[1]{\fnm{Ahmad} \sur{Lotfi}\orcidlink{0000-0001-9028-2563}}\email{ahmad.lofti@ntu.ac.uk}

\author[1]{\fnm{Salisu} \sur{Wada Yahaya}\orcidlink{0000-0002-0394-6112}}\email{salisu.yahaya@ntu.ac.uk}

\author[1]{\fnm{Isibor Kennedy} \sur{Ihianle}\orcidlink{0000-0001-7445-8573}}\email{isibor.ihianle@ntu.ac.uk}

\author[2]{\fnm{Md Mahmudul} \sur{Hasan}\orcidlink{0000-0003-2543-3112}}\email{m.mahmudul@mediprospects.ai}

\author[3]{\fnm{Pedro} \sur{Sousa}\orcidlink{0000-0002-1425-156X}}\email{pedro.sousa@oncontrol-tech.com}

\author*[1]{\fnm{Pedro} \sur{Machado}\orcidlink{0000-0002-1425-156X}}\email{pedro.machado@ntu.ac.uk}

\affil*[1]{\orgdiv{Department of Computer Science}, \orgname{Nottingham Trent University}, \orgaddress{\street{Clifton Campus}, \city{Nottingham}, \postcode{NG11 8NS}, \state{Nottinghamshire}, \country{UK}}}

\affil[2]{\orgname{MediprospectsAI Limited}, \orgaddress{\street{5-7 High Street}, \city{London}, \postcode{E13 0AD}, \country{UK}}}

\affil[3]{\orgname{OnControl}, \orgaddress{\street{Rua Cidade Poitiers, nº 155 – 1º Andar}, \city{Coimbra}, \postcode{3000-108}, \country{Portugal}}}

\abstract{Invasive signal crayfish have a detrimental impact on ecosystems. They spread the fungal-type crayfish plague disease (Aphanomyces astaci) that is lethal to the native white clawed crayfish, the only native crayfish species in Britain. Invasive signal crayfish extensively burrow, causing habitat destruction, erosion of river banks and adverse changes in water quality, while also competing with native species for resources leading to declines in native populations. Moreover, pollution exacerbates the vulnerability of White-clawed crayfish, with their populations declining by over 90\%. To safeguard aquatic ecosystems, it is imperative to address the challenges posed by invasive species and pollution in aquatic ecosystem's. 

This article introduces the \gls*{ced} computing platform for the detection of crayfish and plastic. It also presents two publicly available underwater datasets, annotated with sequences of crayfish and aquatic plastic debris. Four \gls*{yolo} variants were trained and evaluated for crayfish and plastic object detection. \gls*{yolo}v5s achieved the highest detection accuracy, with an mAP@0.5 of 0.90, and achieved the best precision at 0.93. \gls*{yolo}v8s exhibited the best recall at 0.84 and the lowest validation box loss of 0.95. While \gls*{yolo}v5n and \gls*{yolo}v8n also performed well with mAP@0.5 scores of 0.87 and 0.86, respectively, the \gls*{yolo}v5 variants generally showed higher residual errors and notably slower convergence, with \gls*{yolo}v5n being the slowest to stabilise. For real-time embedded testing on the \gls*{njon}, \gls*{yolo}v8n proved to be efficient on the \gls*{gpu}. It recorded the fastest inference time at 0.29 seconds and the lowest energy consumption at 0.09\gls*{j}, \gls*{yolo}v5n took 0.48 seconds and consumed 2.43\gls*{j}. Despite not leading in all individual detection metrics, \gls*{yolo}v8n's balanced mAP@0.5, fastest speed and minimal power draw makes it most suited for deployment on resource-constrained platforms like the \gls*{njon}, where a compromise between accuracy and operational efficiency is required. \gls*{cpu} inference times were notably slower and energy consumption was higher across all models.

The \gls*{ced} platform can play a crucial role in environmental monitoring by performing on-the-fly detection of Signal crayfish and plastic debris while leveraging the efficacy of \gls*{ai}, \gls*{iot} devices and the power of edge computing (i.e., \gls*{njon}). By providing accurate data on the presence, spread and abundance of these species, the platform can contribute to monitoring efforts and aid in mitigating the spread of invasive species.}




\maketitle
\section{Introduction}

Invasive species cause substantial damage to environmental ecosystems globally, this is comparable to the effects of pollutants such as discarded plastic. Extensive information exists about these species, including their characteristics, invaded regions, invasion pathways, and impacts \citep{tekman2022impacts, hobbs2000invasive}. However, there is still a need for comprehensive assessments that consider multiple aspects of the invasion process simultaneously \citep{bernery2022, woods2016towards}.

Invasive signal crayfish (\emph{Pacifastacus Leniusculus}) pose a threat to native species and habitats, causing substantial economic consequences. They out-compete and prey on native white-clawed crayfish (\emph{Austropotamobius Pallipes}), causing population declines and even extinctions \citep{MATHERS2016207}. The white-clawed crayfish is a keystone species found in freshwater bodies in Western Europe \citep{mirimin2022}. However, its populations have experienced significant declines due to the presence of a contagious fungal–type disease (Aphanomyces Astaci) originating from North America, competition with non-indigenous crayfish species, and habitat deterioration \citep{mirimin2022}. Additionally, their burrowing activities result in habitat degradation, destabilising riverbanks and disrupting aquatic vegetation. The economic impact of these damages includes the loss of biodiversity, reduced productivity in fisheries, and the need for habitat restoration, which can require substantial financial investments \citep{FALLER20161190}. Access to accurate and comprehensive data on these invasions is crucial for informed decision-making in environmental management. \citep{bernery2022}. 

\gls*{ml} has revolutionised various fields, including environmental monitoring, by enabling real-time data analysis and decision-making. In aquatic environments, there is an increasing demand for automated systems capable of performing accurate and timely classification of species, pollutants, and underwater infrastructure conditions \citep{Juan_and_Xu}. Traditional methods of aquatic monitoring, which often rely on manual data collection and expert analysis, are time-consuming, resource-intensive, and limited in their scalability. The integration of \gls*{ml}, \gls*{iot} devices, and edge computing offers a promising solution to these challenges, allowing for faster, more efficient, and cost-effective monitoring systems \citep{Goodwin_Morten}.

Use of cameras for biodiversity monitoring has less impact on biodiversity and can cover larger areas and for longer periods compared to traditional observation methods, that involve observers walking in transects in challenging environments. Data that can be collected includes abundance, density, species richness and proportion of area occupied \citep{STEPHENSON202036}. Additionally, various sensors can be added to collect more holistic datasets. For example \gls*{lidar} sensors to measure distance, temperature sensors for temperature measurements and light intensity sensors.

\cite{biber2013} concluded that conventional methods for monitoring invasive species and environmental pollution typically involve manual sampling, which is resource-intensive, slow and unable to provide real-time information required for timely intervention strategies. Thus, environmental conditions monitoring requires innovative approaches that can provide cost-effective, continuous, automated, and scalable monitoring solutions.

To address the challenges of monitoring invasive species and plastics pollution, this paper introduces \gls*{ced}, a novel integration of \gls*{ai}, \gls*{iot}, and edge computing technologies. \gls*{ced} enhances real-time detection of invasive species and plastic debris, providing actionable insights to support conservation efforts and environmental management.

The key contributions in this article are below:
\begin{enumerate}
\item 
A significant bottleneck in developing effective underwater AI solutions is the lack of diverse and well-annotated datasets. Our motivation includes contributing to the research community by providing two publicly available underwater datasets, annotated with sequences of crayfish and aquatic plastic debris. These datasets, which feature challenging underwater conditions, aim to foster further research and development in this critical domain by providing standardised benchmarks for model training and evaluation.
 \item 
Traditional monitoring methods, such as manual sampling and visual surveys, are labor-intensive and prone to observer bias. Traditional monitoring methods are also limited in spatial and temporal resolution, which makes it challenging to track species like the invasive signal crayfish or detect the distribution of plastic debris effectively. \gls*{ced} addresses these limitations by using state-of-the-art deep learning models, specifically \gls*{yolo}-based object detection frameworks, to automate the identification of target species and debris in real time. 

\item Monitoring efforts often rely on centralised systems that collect data remotely and process it later, introducing delays in decision-making. The latency can result in missed opportunities for timely intervention, such as rapid response to invasive species or plastic pollution events. \gls*{ced} integrates edge computing to process data locally, allowing instant analysis and on-the-fly decision-making.

\item 
Manual monitoring programs often incur high costs due to personnel requirements and extensive fieldwork. Additionally, traditional equipment for automated monitoring can be expensive and energy-intensive. \gls*{ced} reduces these costs by utilising lightweight hardware with efficient power consumption, enabling long-term deployment even in remote areas.
\end{enumerate}

Our research aligns with global calls for leveraging technology to combat biodiversity loss and mitigate the impacts of human-induced environmental changes \citep{smith2022}. It offers a significant step forward in integrating \gls*{ai}-driven solutions into ecological monitoring frameworks, bridging gaps between research, policy, and conservation practice.
The rest of the paper is organised as follows. Section \ref{sec:lr} reviews relevant literature. Section \ref{sec:Methodology} outlines the methodology, including an overview of the \gls*{ced} prototype hardware, the datasets used, and the network architecture. Section \ref{sec:results} and \ref{sec:mppc} presents the training and experimental results, with a focus on model training results, stability and power consumption on the \gls*{njon}. Finally, Section \ref{sec:cfw} concludes the paper and discusses directions for future work.

\section{Related Work}\label{sec:lr}

Recent years have seen a significant increase in the interest in processing underwater images \citep{mittal2022}. The interest is driven by the need to study and monitor aquatic plants and animals, which has applications in marine biology, economy, and biodiversity management. Understanding the behaviour and numbers of aquatic species is crucial for various purposes, including protecting endangered species, analysing differences in species, and early detection of climatic events like pollution and global warming. For instance, Plankton, which produces over 80\% of the world's oxygen \citep{witman-2021}, plays a critical role in the ocean's food chain and atmosphere-ocean connection. Monitoring plankton levels is essential, as both low and excessive levels can have harmful consequences. Furthermore, organisms like Posidonia Oceanic, which thrive in clean water, are vital for biodiversity, beach erosion reduction, and water quality enhancement \citep{mittal2022}. Image processing complements other techniques like physiochemical water analysis and sonar-based detection in understanding the impact of global warming and human activity on marine life.

\subsection{The challenges of underwater monitoring}
Underwater imaging contends with substantial energy loss, resulting in diminished and inconsistent illumination and visibility, particularly in the deeper expanses \citep{mittal2022}. The situation is further exacerbated by the presence of freshwater and ocean currents, intensifying the degradation of visibility. Impurities, suspended solids, low contrast, and compromised edges and details further impact underwater images. The distortion of colour based on distance, caused by non-uniform spectral propagation, adds another layer of complexity. Additionally, underwater conditions present intricate and uncontrolled backgrounds, posing challenges in segregating subjects such as sea cucumbers and shellfish from their surroundings. The unbridled movement of fishes in three-dimensional space, often concealed behind objects, compounds the difficulty in accurately determining their pose, orientation, and size. The scarcity of large-scale, publicly available datasets for underwater images, coupled with imbalances, necessitates the incorporation of data augmentation techniques.

The limited availability of training data complicates the training of deep \gls*{cnn}. Moreover, the severe constraints on wired/wireless communication underwater make streaming of underwater imaging exceptionally challenging, resulting in lower-resolution images. This limitation hampers the distinction of features, diminishing the effectiveness of conventional techniques. Conventional \gls*{cv} methods lean on hand-crafted features like \gls*{sift}, \gls*{hog}, and local binary patterns. However, these features lack generalisation capabilities across diverse classes, scenarios, and datasets. The accuracy of traditional techniques plateaus with increasing training data, constraining their overall performance. Extracting hand-crafted features demands domain expertise and time, rendering them less practical for underwater image processing. Previous work on fish recognition often involves dead fish or fish in unnatural conditions, such as swimming pools, limiting its relevance to live marine environments.

\subsection{Early advances in underwater bio-diversity monitoring}
In the early 90's, \cite{strachan-1990} laid the groundwork for automated fish grading based on size and colour, utilising image analysis to estimate fish shape and colour. The proposed approach allowed the differentiation of haddock fish stocks from the Rockall Plateau, located midway between Greenland and Iceland and the North Sea. The method involved digitised \gls*{rgb} images captured from beneath a conveyor belt, processed using bespoke software written in the C programming language. \cite{strachan-1990} marked a significant advancement in distinguishing between fish species. Building on the pioneering work, \cite{storbeck-2001} conducted further research in 2001 by integrating computer vision with a neural network, expanding on the foundations set by \citep{strachan-1990}. Both studies required the fish to be laid flat on a conveyor belt moving at a specific speed, with a vertical arrangement of a laser source and camera above the belt. However, the classification method, while groundbreaking, proves impractical for underwater environments.

\subsection{Modern application of computer vision for underwater bio-diversity monitoring}
A more contemporary perspective is presented in the work of \cite{Zhang}, showcasing a modern application of computer vision for underwater object detection. Employing the \gls*{yolo}v4 model, the authors achieved an \gls*{map} of 92.65\% at a processing speed of approximately 45 \gls*{fps}. Despite the success, it's crucial to note that the achievement relied on a high-end NVIDIA graphics card for accelerating the \gls*{yolo}v4. However, deploying such hardware alongside sensors in physical underwater settings becomes unfeasible due to size and power constraints. In response to the constraint, it is advisable to explore embedded systems as viable alternatives. Such platforms present a more pragmatic deployment option for sensors, ensuring the practical implementation of advanced computer vision techniques in underwater environments. As underwater computer vision emerges as a burgeoning field, studying analogous works becomes imperative to identify potential gaps and opportunities that can be leveraged to advance knowledge in this domain. For instance, \cite{Hegde2021} successfully implemented a trained MobileNetV2 model capable of running on a Raspberry Pi, achieving an impressive precision of 98\% for mackerel. While acknowledging that a \gls*{yolo} model could offer faster results, they deemed it unfeasible due to hardware limitations. Consequently, their image capture and processing occurred at a rate of once every 2 seconds, equivalent to about 0.5 \gls*{fps}. This limitation could prove critical in scenarios involving Underwater Autonomous Vehicles, where objects of interest might traverse the sensors before processing completion.

The incorporation of a \gls*{yolo}v5 model on the \gls*{njon} platform is set to enhance performance, delivering a finely tuned solution tailored for the target applications. Noteworthy is the study by \cite{Hegde2021}, which had a limited implementation of comprehensive image pre-processing methods, mainly focusing on cropping. In contrast, our \gls*{ced} algorithm surpasses these limitations by integrating background subtraction which is a technique enables to extract moving objects (foreground) from the background, as discussed by \cite{Huang2017} and \cite{Kalsotra2021}. 

The sophisticated enhancement is anticipated to streamline object identification, potentially fortifying our algorithm's efficacy in detecting small objects \citep{Hegde2021}. Importantly, our dataset boasts an extensive collection of 450~500 images for each category, significantly contributing to the robustness of our approach. In a comparable study, \cite{Wang} employed \gls*{yolo} nano on an \gls*{njon} to accelerate the algorithm at the edge. The model used by \cite{Wang} demonstrated the capability to detect underwater objects at an impressive frame rate of 8 \gls*{fps}, accompanied by a notable \gls*{map} of approximately 74.8\%. A discernible trade-off emerges between runtime and accuracy when juxtaposed with the work of \cite{Hegde2021}. The strategic use of image pre-processing might provide an avenue for simpler, faster models to attain higher accuracy in underwater object detection scenarios.

\cite{Chen} presented a methodology similar to the one employed in this work, utilising data augmentation to enhance the capabilities of the object detection model. \cite{Wei} proposed mosaic augmentation as use to augment the dataset to enhance the detection of smaller objects in cluttered environments. Moreover \cite{Wei} introduced an innovative adaptation, involving the calculation of gains on the \gls*{rgb} channels to modify and adjust images, accommodating variations in underwater light levels. The \gls*{ced} distinguishes itself through the deliberate exclusion of the background.

Considerable efforts have been directed towards advancing efficient autonomous underwater \glspl*{uav} employing machine learning algorithms for debris detection. Leveraging the computational power of an NVIDIA Jetson TX2 \gls*{gpu}, \cite{fulton-2019} conducted comprehensive experiments involving transfer learning on various models, including \gls*{yolo}v2, Tiny-\gls*{yolo}, Faster R-\gls*{cnn}, and \gls*{ssd}. In their approach, the author fine-tuned each network individually for R-\gls*{cnn}, \gls*{ssd}, and Tiny-\gls*{yolo}. However, for \gls*{yolo}v2, an additional step involved training using transfer learning. The models were specifically tailored for underwater plastic debris detection, and the results indicated that Faster R-\gls*{cnn} achieved the highest accuracy, albeit with longer inference times. \gls*{yolo}v2 struck a balance between speed and accuracy, \gls*{ssd} exhibited the best inference time, and Tiny-\gls*{yolo} emerged as the top performer in terms of overall performance.

\cite{saleh-2024} asserts that automated fish phenotyping, encompassing parameters such as fish length, weight, size, and mortality, provides invaluable insights into understanding the comprehensive profile of classified fish. The task of individually counting fish is notably resource-intensive, demanding skilled labour for hours, thereby contributing to elevated project costs and extended delivery timelines. In a related domain, \cite{ranjan-2023} introduced a real-time mortality monitoring system based on \gls*{yolo}v7, achieving a commendable \gls*{map} of 93.4\% and an F1 score of 0.89. The \gls*{ced} builds upon this foundation, not only incorporating these crucial findings, but also extending the phenotyping metrics. Specifically, we introduce age estimation, distinguishing between adult and juvenile crayfish.

\section{Methodology}\label{sec:Methodology}
The methodology section is divided into four parts. The first part introduces the \gls*{ced} platform workflow, followed by the hardware setup. The third subsection explains the \gls*{yolo}v5n, \gls*{yolo}v5s, \gls*{yolo}v8n and \gls*{yolo}v8s model architectures and final subsection reviews the dataset augmentation techniques applied to the crayfish and underwater plastic datasets. \gls*{yolo} was selected for \gls*{ced} object detection task due to its strong performance in terms of speed, accuracy, and deployment efficiency. In contrast to two-stage detectors like Faster R-CNN~\citep{ren2015faster}, which first generate region proposals and then classify them, \gls*{yolo} uses a single-stage architecture that directly predicts bounding boxes and class probabilities from the full image in one forward pass~\citep{redmon2016you}. These characteristics are particularly important for environmental monitoring applications, where models may need to run on embedded or edge devices with limited computational resources. \gls*{yolo}’s lightweight variants (e.g., \gls*{yolo}v5n, \gls*{yolo}v8n) can be deployed efficiently on such platforms while still providing reliable detection of small, variably shaped targets like fish and discarded plastic debris ~\citep{jocher2020yolov5,glenn2023yolov8}. The source code used in this work is available at \url{https://github.com/denomon/CognitiveEdgeDeviceForRTEMonitoring}.

\subsection{\gls*{ced} prototype overview}
\begin{figure}[h!]
	\centering
	\includegraphics[scale=0.6]{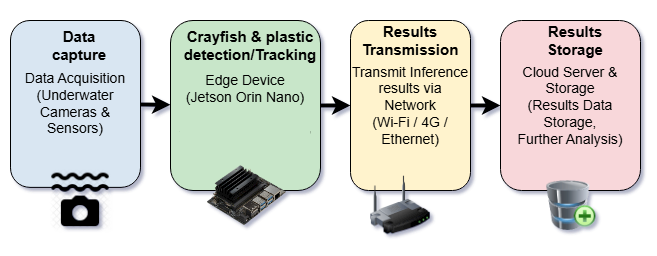}
\caption{\gls*{ced} monitoring workflow - end-to-end process of underwater monitoring. The system begins by recording underwater videos, followed by data import where camera and sensor metadata are stored and video frames extracted. The detection and tracking module then performs real-time identification of Signal Crayfish and plastic waste across frames. Finally, the data archiver formats metadata such as timestamps, geo-location, and depth, storing both detection and sensor data for later export and analysis, enabling biodiversity monitoring and pollution assessment.}
\label{fig:nvidia_jetson_monitoring_system}
\end{figure}

A high level view of the implemented prototype for aquatic monitoring is illustrated in Figure \ref{fig:nvidia_jetson_monitoring_system}. The system features an \gls*{njon} equipped with an Intel RealSense D435i depth sensor. The \gls*{njon} is used for accelerating the \gls*{yolo}v5 and \gls*{yolo}v8 inference at the edge. The \gls*{njon} selected is tailored for diverse tasks, including image identification, object detection, data segmentation, and audio processing. The \gls*{njon} contains a 512-core NVIDIA \gls*{gpu} with 16 Tensor Cores, 6-core Arm Cortex-A78AE v8.2 64-bit \gls*{apu} and 8GB 128-bit LPDDR5 with 68 GB/s bandwidth.

Figure \ref{fig:nvidia_jetson_monitoring_system} below shows four distinct \gls*{ced} workflow modules. The first module, \textbf{Data capture module} is responsible for underwater vision data capture using specialised underwater cameras. The videos serve as the primary inputs for the signal crayfish and plastic detection module and are recorded continuously to ensure comprehensive coverage of the surveyed area. The module is also enhanced with sensors for temperature, salinity, pH, dissolved oxygen, and depth to provide vital context for underwater monitoring. Combining sensor data with vision data allows richer, multi-modal analysis of aquatic environments and biodiversity.

The second module, The \textbf{Signal Crayfish and Plastic Detection \& Tracking module} handles data extraction, processing, signal crayfish and plastic classification and tracking. Data preparation includes image resizing, scaling, and noise reduction. \gls*{yolo} 
 pre-trained \gls*{dl} models perform inference, identifying and tracking Signal crayfish and plastic objects within the extracted frames. The module summarises sensor data, linking detected objects with corresponding timestamps and environmental parameters. The final stage transmitts the results via network to cloud servers for onward storage and analysis. Sensor metadata, detection results, and positional information are stored for analysis whenever required in the future.

\subsection{Network Architecture}
Four object classification network architectures were used from the \gls*{yolo} family used in this research project, driven by the necessity to perform object detection at the edge where processing and power resources are constrained. The choice leaned towards the compact and efficient small and nano versions of \gls*{yolo}. \gls*{yolo}v5n is a lightweight version of the \gls*{yolo}v5 object classification algorithm, \citep{ultralytics}. It was the first classification \gls*{dl} classification architecture and the smallest and fastest model in the \gls*{yolo}v5 family due to striking a balance between model weight and detection accuracy. It has improved detection speed while maintaining a high degree of accuracy compared to previous \gls*{yolo} versions \citep{liu-2023}.

\gls*{yolo}v5s achieves a balance between detection accuracy and speed, stood out as another significant choice. The object classification algorithm contains fewer layers and parameters compared to other \gls*{yolo}v5 models, \gls*{yolo}v5s prioritised efficiency and ease of training, making it suitable for real-time applications. Despite potential sacrifices in detection accuracy compared to larger models, \gls*{yolo}v5s is used in various domains, including liquid biopsy of lung cancer and pouring robot object detection \citep{zhao-2023}.

Moving to the \gls*{yolo}v8n, the smallest and fastest model in the \gls*{yolo}v8 family with 3.2 million parameters, it introduced several architectural improvements over \gls*{yolo}v5 \citep{selcuk2023comparison}. \gls*{yolo}v8n has improved accuracy, faster object detection speed, higher \gls*{map}, and the exclusion of anchor boxes during object detection. The enhanced speed and accuracy made \gls*{yolo}v8n particularly well-suited for real-time detection tasks, aligning seamlessly with the project's objectives \citep{selcuk2023comparison, wang2023uav}.

The fourth model, \gls*{yolo}v8s has more parameters than the other three \gls*{yolo} models used in this article, 11.2 million parameters, 8 million more than \gls*{yolo}v8n. The model also has more \gls*{flops} than its smaller \gls*{yolo}v8n counterpart \citep{selcuk-2023, selcuk2023comparison, wang2023uav} resulting in improved performance than the three \gls*{yolo} models used in this article.

\subsection{Datasets Pre-processing}
\begin{figure}[]
	\centering
	\includegraphics[scale=0.8]{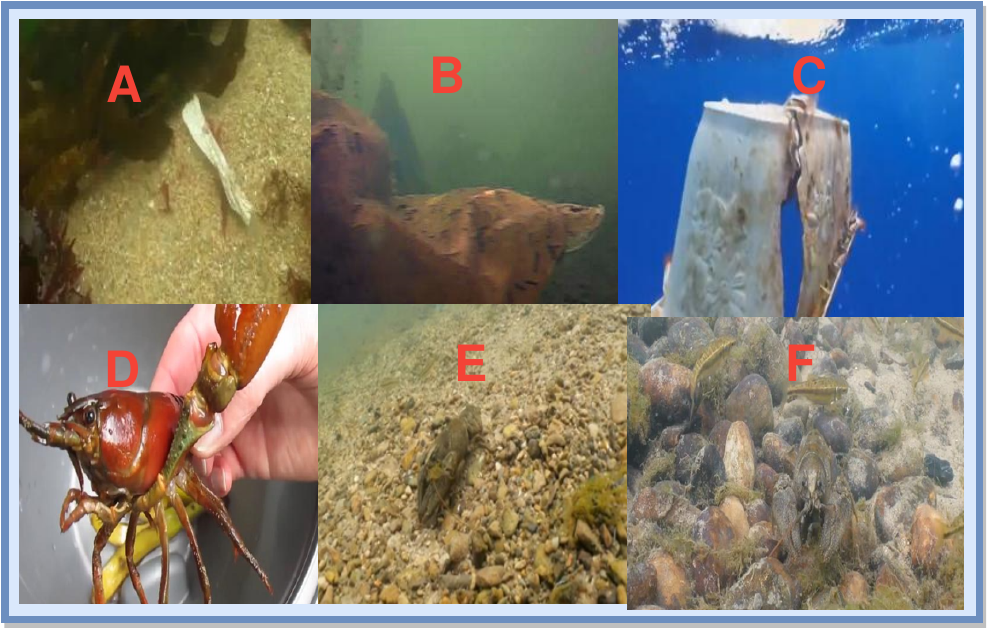}
	\caption{A compilation showcasing a variety of underwater images that are included in the underwater plastic and signal crayfish dataset. 
A) A clear view of a sandy or gravelly bottom. There's a piece of decaying plastic material. 
B) A large, brownish plastic material in deeper or murkier water, obscuring the view causing limited visibility of the surrounding environment.
C) A partially submerged plastic debris 
D) A close-up shot of a signal crayfish being held by a hand with claws and legs clearly visible.
E) A view of a sandy or gravelly bottom with a juvenile signal crayfish partially visible.
F) Numerous juvenile signal crayfish swimming amongst rocks and shells on the bottom.}
 \label{fig:Crayfish_plastic_sample}
\end{figure}

The authors have previously generated a public custom underwater crayfish dataset, consisting of 2486 images \citep{machado_2023}. The proposed dataset was partitioned into three subsets: 1740 images for the training set, 249 for testing, and 497 images for validation. Additionally, another public custom dataset, also created by the authors, focuses on plastic and comprises 1220 images \citep{machado_2022}. Similarly, the plastic dataset was divided into three segments: 854 images for the training set, 122 for testing, and 244 images for validation. A sample is shown in Figure \ref{fig:Crayfish_plastic_sample}.

The dataset contains images captured from real underwater environments that contain crayfish and a wide variety of plastic debris in different stages of decay and are either clearly visible or covered with varying amounts of overgrowth or other underwater debris. Furthermore, lighting quality and water clarity vary greatly across the dataset which makes the images resemble real-world conditions that the \gls*{ced} prototype is intended to work in.

Image labelling is an important step whereby labels or tags are assigned to images based on their content in order to categorise and organise them in a machine-readable format. The Roboflow computer vision platform, \citep{roboflow} was used to accomplish the labelling task. Using Roboflow, the dataset was labelled with two classes, defined as 
\textbf{Crayfish}: all signal crayfish or \textbf{Plastic}: all plastic debris/material.
\subsection{Dataset Augmentation}
Fine-tuning \gls*{yolo} models for optimal performance goes beyond merely implementing the algorithm itself. An essential aspect of the labelling process is the utilisation of image augmentation techniques, such as cropping, flipping, rotating, and colour shifting, which can enhance the model's generalisation and robustness \citep{xu-2018}. Data augmentation, which if not properly customised, may not yield desired model performance. The various data augmentation techniques performed on the crayfish and underwater plastic datasets introduced in this article are discussed below.

Image Scale augmentation is the action of resizing the input images into various dimensions or scales in order to train the \gls*{yolo} model on a dataset containing a variety of scales. This adapts the model to objects of differing sizes, which is the case in real-world scenes. The images in the dataset were stretched to 416x416 to ensure consistency during training and inference.

Images in the dataset were vertically and horizontally flipped. Vertically flipping creates a mirror image where the top becomes the bottom and vice versa. Vertically flipping helps the \gls*{yolo} model detect objects that may appear upside down in real-world scenarios. Similarly, flip left-right augmentation horizontally flips an image, allowing the model to learn and detect objects from different perspectives. Training on these flipped images enhances the model's robustness, making it adaptable to accurately detect objects regardless of their orientation

In cases where the subject is obscured by other objects, only a portion of the object can be captured in the photograph. Therefore, we randomly cropped the images, with both the cropping width and height ranging from 0\% Minimum Zoom, to 49\% Maximum Zoom.

Mosaic augmentation, a method that combines multiple images into a single mosaic-like training sample was also applied to the dataset. Mosaic augmentation technique enhances the \gls*{yolo} model's ability to detect objects in complex scenes with overlapping or crowded environments. By training with mosaic-augmented images, the model becomes more adept at handling situations where objects are partially hidden or blend together, ultimately improving its accuracy in challenging scenarios.

The hue component of images in the dataset was altered in the range between -25° and +25°, saturation altered in the range between -42\% and +42\%, exposure altered between -22\% and +22\% and Gray-scale applied to 47\% of images. Adjusting the hue component mimics diverse lighting conditions, including daylight or artificial lighting. Altering the hue component allows the model to train and recognise objects effectively under a range of illumination settings Applying the Hue, Saturation and exposure augmentation ensures the trained model is robust enough to handle real world lighting exposure, various colour schemes and contrasts

\section{Result Analysis}\label{sec:results}
The results for training, testing and validation are discussed in this section. In the previous section, a classifier was trained with 2594 images in the training set and evaluated its performance on 371 images in the test set. The classifier's accuracy was assessed based on three measures: training accuracy, test accuracy, and validation accuracy. Training accuracy refers to how accurately the classifier performed on the training set, while test accuracy measures its performance on the test set. Validation accuracy indicates the rate at which the classifier successfully classified newly encountered images in the validation set. Additionally, we considered the recorded losses to evaluate the effectiveness of our approach.

The models will be evaluated according to the highest \gls*{map}, F1 score and average precision. The \gls*{map}, weighted mean of precision's at each threshold, (i.e a comparison score between the ground-truth bounding box to the actual detected box), is useful where many classes are present in object detection with each class having differing detection performance. A higher \gls*{map} score indicates a higher detection accuracy.
The formulae for \gls*{ap} is rulled by Equation \ref{eq:mean_average_precision} below. \gls*{map} is the mean of AP over all the queries.

\begin{eqnarray}
\label{eq:mean_average_precision}
\text{\gls*{ap}} = \sum_n (R_n - R_{n-1}) P_n
\end{eqnarray}

\noindent where $R_n$ and $P_n$ are the precision and recall at the $n$th threshold.

Two functions that return an array of labels were created to produce 18 different confusion matrices. 
The first function grabs all the true values from the test set. The second function grabs all the predicted labels from each model with the highest \gls*{map} value per augmentation. The two arrays of values are then used to plot the confusion matrices. The confusion matrix, which summarises the counts of \gls*{tp}, \gls*{tn}, \gls*{fp}, and \gls*{fn}, provides the necessary values for calculating recall (\ref{eq:recall}), precision (\ref{eq:precision}), and F1-score (\ref{eq:f1}).

\begin{eqnarray}
\label{eq:recall}
Recall: re =\frac{tp}{tp+fn}
\end{eqnarray}

\begin{eqnarray}
\label{eq:precision}
Precision: pr =\frac{tp}{tp+fp}
\end{eqnarray}

\begin{eqnarray}
\label{eq:f1}
F1-score: f1 = 2\times \frac{pr\times re}{pr+re}
\end{eqnarray}

\noindent where, $tp$ is the true positives, $tn$ is the true negatives, $fp$ is the false positives and $fn$ is the false negatives.

\subsection{Training Results}
In this article, the \gls*{yolo}v5n, \gls*{yolo}v5s, \gls*{yolo}v8n and \gls*{yolo}v8s model architectures are trained on the Underwater plastic and crayfish datasets. After training, the four models are applied to the test set of the Underwater plastic and crayfish datasets and measured results of crayfish and plastic detection are obtained. 

The figure \ref{fig:yolo_combined_pr_curve} below presents the precision-recall curve scores obtained during model training. The graphs all show a high area under the curve, which signifies a combination of high recall and high precision. High precision indicates that the model achieved a low false positive rate, which means that the classifier accurately identified crayfish without many false positives. however, high recall corresponds to a low false negative rate, implying that the classifier successfully captures the majority of positive results. When both precision and recall scores are high, it indicates that the classifier is returning accurate and relevant results (high precision) while also capturing a significant portion of all positive results (high recall). Consequently, an analysis of the results of our training indicate that \gls*{yolo}v5s achieved the highest precision among the four \gls*{yolo} architectures achieving a score of 0.93 followed by \gls*{yolo}v8s that achieved a score of 0.91. \gls*{yolo}v8n achieved a score of 0.90 and finally \gls*{yolo}v5n achieving an precision score of 0.88. These results are shown in table \ref{table:yolo_vaildation_scores} and figures \ref{fig:yolo_combined_pr_curve} and \ref{fig:yolo_combined_f1_score} below. 

\begin{figure}[]
	\centering
	\includegraphics[scale=0.6]{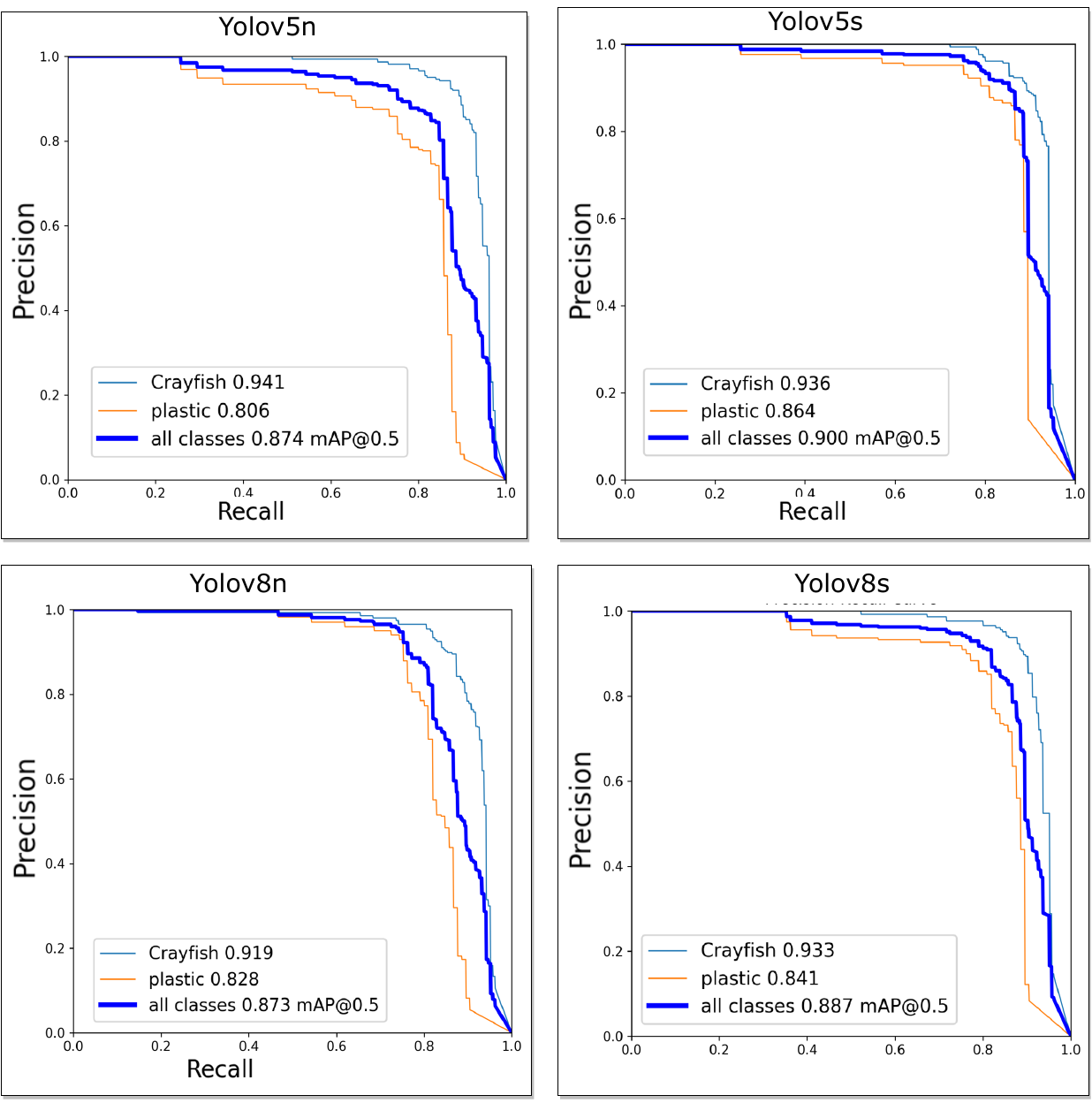}
	\caption{\gls*{pr} curves for \gls*{yolo}v5n, \gls*{yolo}v5s, \gls*{yolo}v8n and \gls*{yolo}v8s models on crayfish and plastic detection. The light blue and orange curves represent the crayfish and plastic classes while the bold blue curve indicates the overall \gls*{map} at IoU 0.5. \gls*{yolo}v5s achieves the highest \gls*{map} (0.900), followed closely by \gls*{yolo}v8s (0.887). }
 \label{fig:yolo_combined_pr_curve}
\end{figure}

Figures \ref{fig:yolo_combined_f1_score} below presents the F1 confidence scores obtained during the training exercise. \gls*{yolo}v5s and \gls*{yolo}v8s achieved the highest scores at F1 score of 0.87. This was followed by \gls*{yolo}v8n that had an F1 score of 0.85 and finally \gls*{yolo}v5s achieving the lowest score of 0.84. 
A summary of these results is also presented in table \ref{table:yolo_vaildation_scores}.

\begin{figure}[]
	\centering
	\includegraphics[scale=0.6]{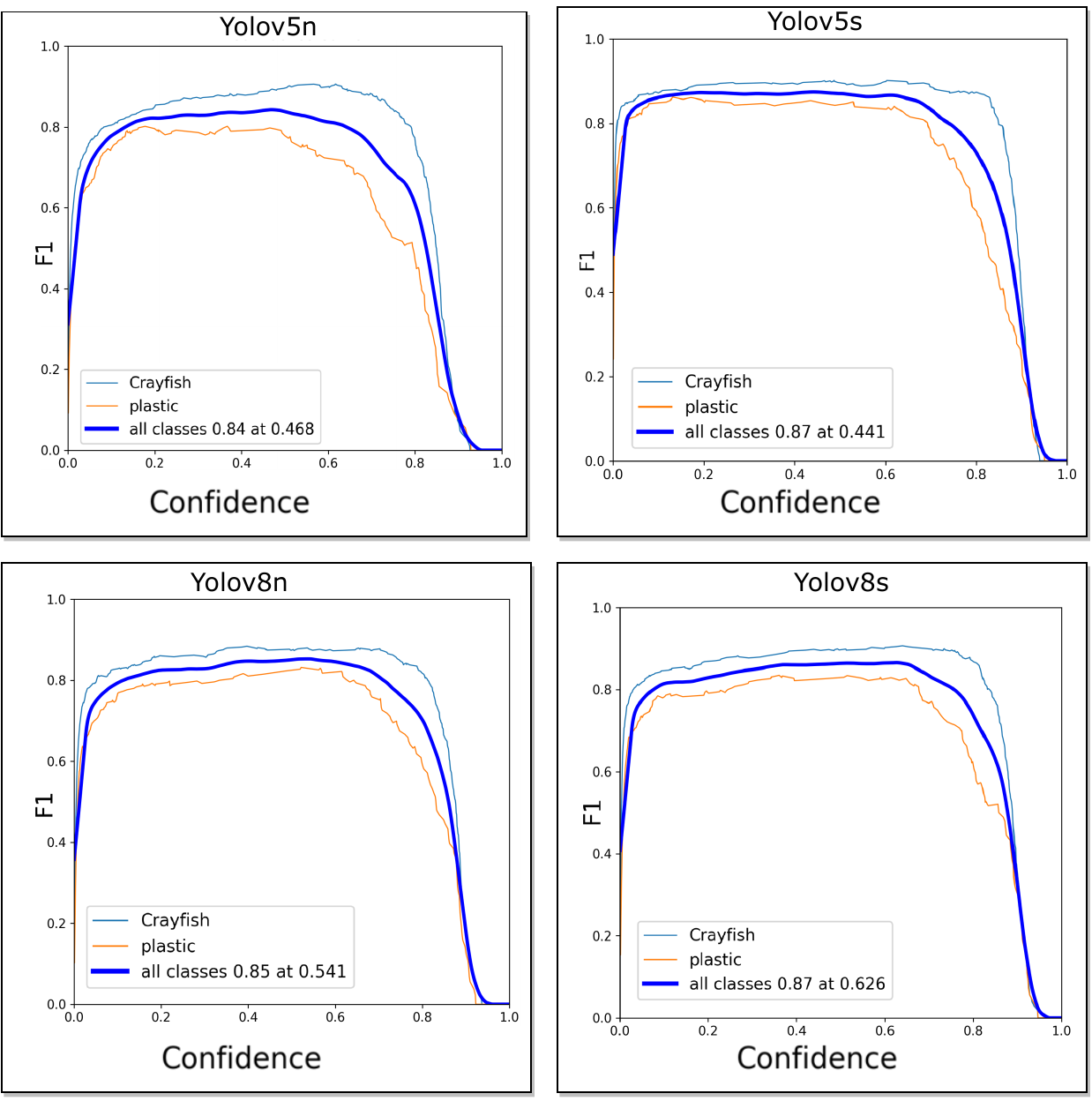}
	\caption{F1-score vs. confidence threshold curves for \gls*{yolo}v5n, \gls*{yolo}v5s, \gls*{yolo}v8n, \gls*{yolo}v8s models. The light blue and orange curves represent the crayfish and plastic classes while the bold blue curve indicates the overall F1-score. \gls*{yolo}v8s achieves the highest F1-score of 0.87 at a confidence threshold of 0.626, followed closely by \gls*{yolo}v8n 0.85 at a confidence threshold of 0.541.}
 \label{fig:yolo_combined_f1_score}
\end{figure}

\begin{table}[]
\centering
\caption{
Comparison of final validation performance across four \gls*{yolo} object‐detection variants on our crayfish and aquatic plastic dataset. Bold indicates the best performance in each column.
}
\begin{tabular}{|c|c|c|c|c|}
\hline
Model & Precision & Recall & F$_1$ Score & mAP@0.5 \\ 
\hline
 \gls*{yolo}v5s & \textbf{0.93} & 0.83 & \textbf{0.87} & \textbf{0.90} \\ 
\hline
 \gls*{yolo}v5n & 0.88 & 0.81 & 0.84 & 0.87 \\ 
\hline
 \gls*{yolo}v8s & 0.91 & \textbf{0.84} & 0.87 & 0.88 \\ 
\hline
 \gls*{yolo}v8n & 0.90 & 0.80 & 0.85 & 0.86 \\ 
\hline
\end{tabular}
\label{table:yolo_vaildation_scores}
\end{table}

\subsection{Convergence analysis}
We evaluated the convergence stability of \gls*{yolo}v5s, \gls*{yolo}v5n, \gls*{yolo}v8s, and \gls*{yolo}v8n based on mAP@0.5 and validation box loss over 580 training epochs. \gls*{yolo}v5s achieved the highest final mAP@0.5 (0.900), followed closely by \gls*{yolo}v8s (0.887). \gls*{yolo}v5n and \gls*{yolo}v8n showed comparable but slightly lower final mAP values (0.874 and 0.873, respectively). The small models (\gls*{yolo}v5s and \gls*{yolo}v8s) converged more quickly and smoothly, with less fluctuation in performance across epochs.

The loss trends of the validation boxes aligned with the mAP curves, \gls*{yolo}v8s recorded the lowest and most stable validation loss, indicating efficient localisation, while \gls*{yolo}v5n exhibited higher and more erratic loss, consistent with its slower and less stable convergence. Overall, \gls*{yolo}v5s demonstrated the best balance between high accuracy and stable training, making it the most effective model for the \gls*{ced} detection task, with \gls*{yolo}v8s also performing robustly. The results are shown in figure \ref{fig:convergence_analysis} below.

\begin{figure}[]
	\centering
	\includegraphics[scale=0.6]{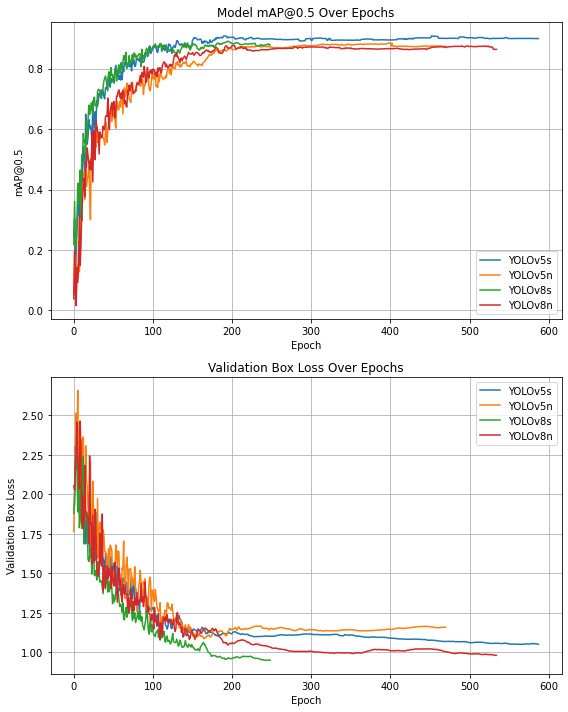}
	\caption{Convergence analysis of \gls*{yolo}v5s, \gls*{yolo}v5n, \gls*{yolo}v8s, and \gls*{yolo}v8n over 580 training epochs. The top plot shows mAP@0.5 progression, where \gls*{yolo}v5s converges fastest and achieves the highest detection accuracy. The bottom plot shows validation box loss, with \gls*{yolo}v8s exhibiting the lowest and most stable loss, indicating superior localization performance. Overall, \gls*{yolo}v8s demonstrates the best convergence speed and training stability among the evaluated models.}
 \label{fig:convergence_analysis}
\end{figure}

\section{Model performance and power consumption on the \gls*{njon}}\label{sec:mppc}
Measuring power consumption and inference time is crucial to ensure that device inference speed and power consumption requirements are adequately addressed. Effective power source planning is essential as it enables the implementation of power backup measures for devices operating remotely at the edge. Such proactive measures aim to minimise the potential risk of downtime caused by power outages. To measure power consumption on the the \gls*{njon}, Jetson-stats, \cite{jetsonstats} python library was used to analyse the average power consumed by the \gls*{njon} platform while performing inference using the four \gls*{yolo} classification models, one at a time. \gls*{cpu}, \gls*{gpu} Inference time and power consumption measurements were the key metrics recorded and the results shown in table \ref{tab:inference_power} are discussed below. Energy consumption was estimated by multiplying the inference time by the change in total board power (VDD\_IN), recorded immediately before and after inference using Jetson-stats. This approach provides a practical approximation for comparing model energy efficiency on the Jetson platform because we are able to capture the change in power consumption during the inference period. We report energy consumption rather than power consumption, as model inference times vary significantly. This provides a fairer comparison of overall resource efficiency on the \gls*{njon}. The \gls*{njon} platform was selected to evaluate its potential for real-time environmental monitoring, especially in aquatic ecosystems where accurate classification of invasive species is critical to the conservation of biodiversity. The results are discussed in sections \ref{subsec:inference_time} and \ref{subsec:power_consumption} below.

\begin{table}[]
\centering
 \caption{Inference time and power consumption for four \gls*{yolo} models on \gls*{cpu} and \gls*{gpu}. \gls*{yolo}v8n highlighted in green colour below achieved the best inference time (fastest) and consumed the least amount of energy per inference on the \gls*{gpu}. \gls*{gpu} consistently achieves best inference time and required lower energy per inference compared to \gls*{cpu}.}
\begin{tabular}{|l|c|c|c|c|}
\hline
\textbf{Model} & \textbf{Device} & \textbf{Inference Time(s)} & \textbf{Power(W)} & \textbf{Energy(J)} \\
\hline
\rowcolor{green!30}\gls*{yolo}v8n & \gls*{gpu} & \textbf{0.2914} & \textbf{0.32} & \textbf{0.093248} \\
 \gls*{yolo}v8s & \gls*{gpu} & 0.3748 & 1.19 & 0.446012 \\
\gls*{yolo}v5s & \gls*{gpu} & 0.4342 & 0.44 & 0.191048 \\
\gls*{yolo}v5n & \gls*{gpu} & 0.4829 & 0.83 & 0.400807 \\
\gls*{yolo}v8n & \gls*{cpu} & 0.5947 & 1.24 & 0.737428 \\
\gls*{yolo}v5n & \gls*{cpu} & 0.8579 & 2.43 & 2.084697 \\
\gls*{yolo}v5s & \gls*{cpu} & 1.3083 & 1.03 & 1.347549 \\
\gls*{yolo}v8s & \gls*{cpu} & 1.3125 & 0.87 & 1.141875 \\
\hline
\end{tabular}
\label{tab:inference_power}
\end{table}

\subsection{Inference Time}\label{subsec:inference_time}
The four \gls*{yolo}-based models showed significant differences in their inference times depending on the device (\gls*{cpu} vs. \gls*{gpu}) used as shown in figure \ref{fig:Jetson_inference_time} below. When executed on the \gls*{gpu}, \gls*{yolo}v8n demonstrated the fastest inference time, completing classification in 0.29 seconds. Conversely, \gls*{yolo}v5n had the longest inference time on the \gls*{gpu}, requiring 0.48 seconds. In contrast, \gls*{cpu} executions were notably slower across all four models. The \gls*{yolo}v8n model took 0.59 seconds on the \gls*{cpu}, while \gls*{yolo}v8s was the slowest with a \gls*{cpu} inference time of 1.31 seconds. These results highlight the efficiency of the \gls*{gpu} in handling real-time object detection tasks, particularly for embedded systems like the \gls*{njon}. Based on these findings, \gls*{yolo}v8n proved to be the fastest model for real-time inference on the \gls*{njon} platform. 

\begin{figure}[]
\begin{center} 
\includegraphics[width=1\linewidth]{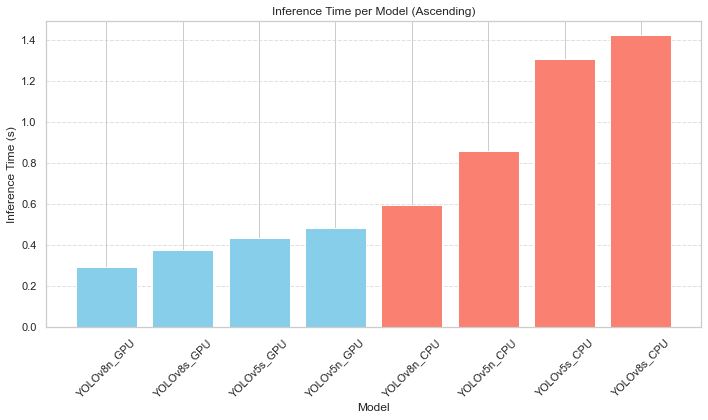}
\end{center}
 \caption{Bar chart showing the inference time of four \gls*{yolo} models on \gls*{gpu} and \gls*{cpu}, ordered by the lowest inference time. The results show that \gls*{gpu} achieves faster inference time compared to the \gls*{cpu} with \gls*{yolo}v8n achieving the lowest results on the \gls*{gpu} and \gls*{yolo}v8s taking longest inference time \gls*{cpu}}
 \label{fig:Jetson_inference_time}
\end{figure}

\subsection{Energy Consumption Measurements}\label{subsec:power_consumption}

To assess the energy efficiency of the deployed \gls*{yolo} models on the NVIDIA Jetson platform, we estimated the energy consumed during inference using power readings obtained via Jetson-stats. The total \gls*{njon} board power consumption was monitored using the \texttt{VDD\_IN} values reported before and after inference.
The increase in power consumption during inference was computed as:

\begin{equation}
\Delta \text{VDD}_{\text{IN}} = \text{VDD}_{\text{IN, After}} - \text{VDD}_{\text{IN, Before}}
\end{equation}

This method provides a consistent way to compare the relative energy costs of different models, under the assumption that the primary power increase during the short inference window is attributable to the model execution.The inference energy consumption \( E \) (in \gls*{j}) was then estimated using:

\begin{equation}
E = \Delta \text{VDD}_{\text{IN}} \times \text{Inference Time (s)}
\end{equation}

From the results shown in figure \ref{fig:gpu_power_consumption} below, it was observed that the \gls*{gpu} generally required less energy than the \gls*{cpu} across all four custom trained \gls*{yolo} models. The \gls*{yolo}v8n model, which had the fastest inference time of 0.29 seconds on the \gls*{gpu}, also demonstrated the lowest \gls*{gpu} energy consumption at 0.09\gls*{j} and \gls*{yolo}v8s had the highest \gls*{gpu} energy consumption at 0.45J per inference. On the \gls*{cpu}, energy consumption ranged between 0.74J and 2.08\gls*{j}, with \gls*{yolo}v8n consuming the least \gls*{cpu} power of 0.74J and \gls*{yolo}v5n consuming the highest amount of power at 2.08\gls*{j}. These findings indicate that \gls*{yolo}v8n is best suited for power-limited edge environments, as it consumes the least amount of power among the four \gls*{yolo} variants.

The results also indicate that \gls*{gpu} execution provides significant advantages in terms of both inference time and power consumption for deep learning tasks like signal crayfish and plastic classification. The \gls*{yolo}v8n model, which exhibited the best balance of performance and low power consumption, is the most suitable option for real-time deployment on the \gls*{njon} platform. On the other hand, \gls*{cpu} execution, while offering flexibility in deployment scenarios, generally results in higher power consumption and longer inference time. These findings are crucial for optimising the performance of edge devices in resource-constrained environments, where both power and speed are essential for practical applications.

\begin{figure}[!htb]
\begin{center} 
\includegraphics[width=1\linewidth]{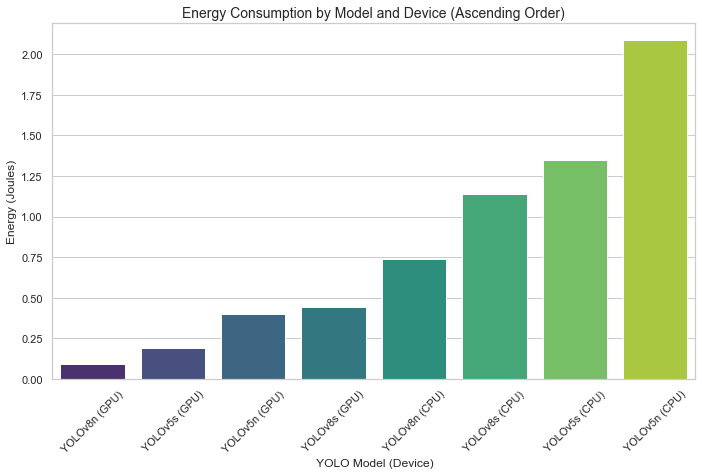}
\end{center}
 \caption{\gls*{njon} \textbf{\gls*{gpu}} and \textbf{\gls*{cpu}} power consumption measurements while performing a classification task using four pre-trained \gls*{yolo} models. \gls*{gpu} power consumption is significantly lower than \gls*{cpu} power consumption with \gls*{yolo}v8n consuming the least amount at 0.09\gls*{j} on the \gls*{gpu} and \gls*{yolo}v5n consuming the most at 2.08\gls*{j} on the \gls*{cpu}}
 \label{fig:gpu_power_consumption}
\end{figure}

\subsection{Stability}
To evaluate the performance of different \gls*{yolo} models, we conducted a series of tests to evaluate the inference time, power consumption, and robustness. The models tested are \gls*{yolo}v5n, \gls*{yolo}v5s, \gls*{yolo}v8n, and \gls*{yolo}v8s, with a focus on three key areas: stability, robustness to Gaussian noise and robustness to motion blur which are common challenges facing underwater environments. The tests were executed on the \gls*{gpu} due to prior tests that confirmed \gls*{gpu} execution provides significant advantages in terms of both inference time and power consumption for deep learning tasks. 

Stability testing involved evaluating the consistency of inference time, power consumption, and confidence stability when the same input image was used repeatedly. The stability of the models was measured by calculating the mean and standard deviation of inference times and power consumption over several runs. \gls*{yolo}v5n exhibited a mean inference time of 0.375 seconds with a relatively high standard deviation (0.495), indicating some variability. \gls*{yolo}v5s showed a higher mean inference time of 0.577 seconds, with even more significant variability (0.684). \gls*{yolo}v8n was the fastest, with a mean inference time of 0.250 seconds and a moderate standard deviation of 0.532. \gls*{yolo}v8s had a mean inference time of 0.459 seconds, with a standard deviation of 0.695, slightly more variability than \gls*{yolo}v8n.

\subsection{Robustness to Gaussian noise and motion blur}

For robustness, the performance of each of the four \gls*{yolo} models was tested under two conditions: Gaussian noise and motion blur. Gaussian noise is commonly used to simulate sensor noise and environmental distortions in underwater imaging, which can arise due to low-light conditions or electronic interference \citep*{zhangWang2021}. Motion blur mimics the effects of fish movement and camera instability, helping assess the robustness of the \gls*{yolo}models in dynamic aquatic environments \citep*{chen2020}. The robustness test with Gaussian noise simulates poor lighting or sensor artifacts in real-world scenarios. \gls*{yolo}v5n showed a slight increase in inference time (0.394 seconds), with a high standard deviation (0.671), but power consumption remained minimal. \gls*{yolo}v5s exhibited a significant drop in mean inference time (0.251 seconds), but its standard deviation increased considerably (0.501), indicating some instability. \gls*{yolo}v8n performed well, with the fastest inference time (0.228 seconds) and a low deviation (0.496), showcasing strong robustness. \gls*{yolo}v8s, on the other hand, was the slowest at 0.594 seconds, with a very high standard deviation (1.065), suggesting instability under noisy conditions.

In the motion blur test, which simulates detection in dynamic settings, \gls*{yolo}v5n showed a decrease in inference time (0.206 seconds) but had a high deviation (0.470). \gls*{yolo}v5s performed inconsistently, with a high inference time (0.600 seconds) and significant variability (1.366). \gls*{yolo}v8n performed exceptionally well, with the lowest mean inference time of 0.056 seconds and minimal variation (0.045), making it highly robust to motion blur. \gls*{yolo}v8s had a high mean inference time of 0.772 seconds, along with a high standard deviation of 0.859, indicating less robustness in this scenario. The table below summarises the results of stability and robustness to noise (Gaussian and motion blur), highlighting the mean and standard deviation of inference times and power consumption for each model and test scenario.

\begin{table}[]
 \centering
 \adjustbox{max width=\textwidth}{
 \begin{tabular}{|p{2cm}|p{5.5cm}|p{2cm}|p{2cm}|p{2.5cm}|p{2cm}|}
 \hline
 Model & Test Type & Mean Inference Time (s) & Mean \gls*{gpu}+\gls*{cpu} Power (W) & Mean \gls*{soc} Power (W) & Mean inference Confidence \\
 \hline
 \gls*{yolo}v5n & Stability & \textit{0.2210} & \textit{0.0676} & \textit{0.0314} & \textit{0.7993} \\
 \gls*{yolo}v5n & Robustness (Gaussian Noise) & \textbf{0.0687} & 0.0160 & 0.0160 & \textit{0.6803} \\
 \gls*{yolo}v5n & Robustness (Motion Blur) & \textbf{0.0719} & 0.0161 & \textit{0.0201} & \textit{0.5265} \\
 \hline
 \gls*{yolo}v5s & Stability & 0.1310 & 0.0325 & 0.0202 & \textbf{0.8728} \\
 \gls*{yolo}v5s & Robustness (Gaussian Noise) & 0.0894 & 0.0285 & 0.0081 & \textbf{0.8579} \\
 \gls*{yolo}v5s & Robustness (Motion Blur) & 0.0912 & 0.0161 & \textit{0.0201} & \textbf{0.8454} \\
 \hline
 \gls*{yolo}v8n & Stability & \textbf{0.0928} & \textbf{0.0119} & \textbf{0.0160} & 0.8448 \\
 \gls*{yolo}v8n & Robustness (Gaussian Noise) & 0.0720 & \textbf{0.0081} & \textbf{0.0001} & 0.7757 \\
 \gls*{yolo}v8n & Robustness (Motion Blur) & 0.0767 & \textbf{0.0160} & \textbf{0.0081} & 0.7024 \\
 \hline
 \gls*{yolo}v8s & Stability & 0.1226 & 0.0647 & 0.0200 & 0.8542 \\
 \gls*{yolo}v8s & Robustness (Gaussian Noise) & \textit{0.0905} & \textit{0.0325} & 0.0080 & 0.7128 \\
 \gls*{yolo}v8s & Robustness (Motion Blur) & \textit{0.0944} & \textit{0.0244} & 0.0121 & 0.8041 \\
 \hline
 \end{tabular}
 }
 \caption{Performance evaluation of different \gls*{yolo} models under various test conditions ( mean inference time and power consumption (watts), and mean confidence). Best values are highlighted in \textbf{bold}, worst in \textit{italics}.}
 \label{tab:yolo_performance}
\end{table}

\section{Conclusions and Future Work}\label{sec:cfw}
The Underwater \gls*{dl} edge computing platform (\gls*{ced}), a novel monitoring platform for underwater crayfish and plastics identification that is world-leading in terms of originality, significance and rigour is presented in this article. The platform integrates diverse methods, including background subtraction for selectively cropping regions of interest in high-resolution images to detect crayfish and underwater plastic debris. By providing accurate data on the presence, distribution, and population density of these species, the \gls*{ced} platform makes a substantial contribution to monitoring initiatives and aids in addressing the spread of invasive species
A public custom underwater crayfish dataset, consisting of 2486 images \citep{machado_2023} was generated as part of this Research Project and made publicly available. Additionally, a second public custom dataset, also created by the authors, focuses on plastic and comprises 1220 images \citep{machado_2022} was generated and released into the public domain.

\gls*{yolo}v5s demonstrated the highest overall detection accuracy (mAP@0.5 of 0.90) and precision (0.93), demonstrating its strong capability to identify crayfish and plastic debris with minimal false positives. \gls*{yolo}v8s achieved the best recall of 0.84 and localisation precision, achieving the lowest validation box loss of 0.95. Although \gls*{yolo}v5n and \gls*{yolo}v8n also performed well with mAP@0.5 scores of 0.87 and 0.86 respectively, the \gls*{yolo}v5 variants generally showed higher residual errors and notably slower convergence, with \gls*{yolo}v5n being the slowest to stabilise. For applications on resource-constrained embedded systems like the \gls*{njon}, \gls*{yolo}v8n emerged as the most optimal model when considering a balance between performance and operational efficiency. On the \gls*{gpu}, it recorded the fastest overall inference time at 0.29 seconds and the lowest total energy consumption at just 0.09 \gls*{njon}, significantly outperforming alternatives like \gls*{yolo}v5n (0.4829 seconds and 0.40 \gls*{j}). Furthermore, \gls*{yolo}v8n consistently exhibited exceptionally low Mean SOC power across various test conditions, reaching as low as 0.0001 W under Gaussian Noise, further underscoring its hardware-level power optimisation. \gls*{cpu} inference times were notably slower across all models (e.g., \gls*{yolo}v8n at 0.59 seconds vs. \gls*{yolo}v8s at 1.31 seconds), reinforcing the necessity of \gls*{gpu} acceleration for real-time applications. In summary, while \gls*{yolo}v5s excelled in overall detection accuracy and precision, and \gls*{yolo}v8s offered strong recall and precise localisation, \gls*{yolo}v8n is identified as the most suitable model for real-time embedded applications. Its efficiency, highlighted by its leading \gls*{gpu} inference speed and lowest energy consumption, combined with its robust performance and excellent SOC power management, directly addresses the demanding operational requirements of underwater autonomous systems.

Future work will focus on equipping \gls*{ced} with pose estimation capabilities to enable detailed analysis of fish behaviour and movement patterns, which can serve as indicators of health and stress. Abnormal swimming patterns, changes in fin position, or irregular body curvature detected through pose estimation may signal injuries, disease, or environmental distress. When combined with contextual sensor data such as water quality, temperature, and oxygen levels, this integrated approach will enhance \gls*{ced}'s ability to monitor aquatic animal health and behaviour in real time, supporting conservation, research, and early intervention strategies. The \gls*{ced} platform will empower policymakers, conservation authorities, and researchers with real-time data on the spread of invasive non-native species and discarded plastic debris floating along water bodies. The availability of such data will facilitate the enactment of well-informed conservation laws and the implementation of effective mitigation measures to safeguard native endangered species and reduce plastic pollution.

While the current study concentrated on identifying discarded plastic and signal crayfish as the non-native species responsible for the decimation of native white-clawed crayfish, the \gls*{ced} platform has the potential to be further trained to recognise other non-native species. For instance, it could be extended to identify invasive non-native species such as the red-eared terrapin and Chinese mitten crab, broadening the scope of its applicability and environmental impact assessment.

\section*{Author Contributions}
D.M. led the implementation of the Cognitive Edge Device (CED) platform, developed the machine learning algorithms, performed the experiments, and conducted the results analysis. F.F.T. contributed to system development, supported the implementation of the detection pipeline, and assisted in analysing and interpreting the experimental outcomes. J.J.B. provided methodological guidance, contributed to refining the machine learning workflow, and supported the review of experimental results. A.L. contributed to conceptual oversight, provided feedback on system design and methodology, and participated in reviewing and improving the manuscript. S.W.Y. assisted with data preparation, supported the experimental setup, and contributed to reviewing the manuscript for technical accuracy. M.M.H. contributed to the interpretation of results, supported the preparation of figures and tables, and provided critical revisions to the manuscript. P.S. contributed to hardware integration, including sensor and embedded platform configuration, and supported the evaluation of system performance in edge deployment scenarios. P.M. conceived and supervised the study, guided the development of the CED platform, contributed to the research design, assisted in results analysis, and led the writing and revision of the manuscript.

All authors reviewed and approved the final manuscript.

\section*{Funding statement}
This research was supported by QR funding from Nottingham Trent University's Department of Computer Science

\section*{Conflict of Interest}
The authors declare that there are no known competing financial interests or personal relationships that could have appeared to influence the work reported in this article.

\section*{Data Availability}
The source code developed for the \gls*{ced} platform, including scripts for dataset preprocessing, model training, inference, and edge deployment, is openly available at \url{https://github.com/denomon/CognitiveEdgeDeviceForRTEMonitoring}. The underwater datasets used in this study are publicly accessible. The crayfish dataset comprising 2,486 annotated underwater images is available on Zenodo at \cite{machado_2023} and the underwater plastic debris dataset, consisting of 1,220 annotated images, is available on Zenodo at \cite{machado_2022}. All datasets and source code used in this work are freely accessible for replication, benchmarking, and further research.

\section*{Clinical trial number}
Not applicable.

\section*{Consent to Publish declaration} 
Not applicable.

\section*{Ethics and Consent to Participate declarations}
Not applicable.

\bibliography{references}
\end{document}

%% file: glossary.tex
\newacronym{dnn}{DNN}{Deep Neural Network}
\newacronym{dl}{DL}{Deep Learning}
\newacronym{ai}{AI}{Artificial Intelligence}
\newacronym{yolo}{YOLO}{You Only Look Once}
\newacronym{gpu}{GPU}{Graphics Processing Unit}
\newacronym{cpu}{CPU}{Central Processing Unit}
\newacronym{soc}{SOC}{System On Chip}
\newacronym{bs}{BS}{Background Subtraction}
\newacronym{rcnn}{RCNN}{Region-based Convolution Neural Network}
\newacronym{cnn}{CNN}{Convolutional Neural Network}
\newacronym{roi}{RoI}{Region of Interest}
\newacronym{njon}{NJON}{NVIDIA Jetson Orin Nano}
\newacronym{fps}{FPS}{Frames per Second}
\newacronym{nics}{NICS}{non-indigenous crayfish species}
\newacronym{map}{mAP}{mean Average Precision}
\newacronym{rgb}{RGB}{Red, Green, and Blue}
\newacronym{uav}{UAV}{Underwater Autonomous Vehicle}
\newacronym{ssd}{SSD}{Single Shot Detector}
\newacronym{lidar}{LIDAR}{Light Detection and Ranging}
\newacronym{apu}{APU}{Application Processor Unit}
\newacronym{cv}{CV}{Computer Vision}
\newacronym{sift}{SIFT}{Scale-Invariant Feature Transform}
\newacronym{hog}{HOG}{Histogram of Oriented Gradients}
\newacronym{flops}{FLOPS}{Floating Point Operations Per Second}
\newacronym{gmm}{GMM}{Gaussian Mixture Model}
\newacronym{ap}{AP}{Average Precision}
\newacronym{csv}{CSV}{Comma Separated Values}
\newacronym{iou}{IoU}{Intersection over Union}
\newacronym{iot}{IOT}{Internet of Things}
\newacronym{uk}{UK}{United Kingdom}
\newacronym{ced}{CED}{Cognitive Edge Device}
\newacronym{pr}{PR}{Precision-Recall}
\newacronym{j}{J}{Joules}
\newacronym{ml}{ML}{Machine Learning}
\newacronym{tp}{TP}{True Positives}
\newacronym{tn}{TN}{True Negatives}
\newacronym{fp}{FP}{False Positives}
\newacronym{fn}{FN}{False Negatives}

%% file: references.bib
@article{bernery2022,
  author    = {Bernery, C. and Bellard, C. and Courchamp, F. and Brosse, S. and Gozlan, R. E. and Jarić, I. and Teletchea, F. and Leroy, B.},
  title     = {Freshwater fish invasions: A comprehensive review},
  journal   = {Annual Review of Ecology, Evolution, and Systematics},
  volume    = {53},
  pages     = {427--456},
  year      = {2022},
  url       = {https://hal.science/hal-03781186v1/file/MS_REVIEW_HAL.pdf}
}

@article{mirimin2022,
  title = {Investigation of the First Recent Crayfish Plague Outbreak in Ireland and its Subsequent Spread in the Bruskey River and Surrounding Areas},
  author = {Mirimin, Luca and Brady, Daniel and Gammell, Martin and Lally, Heather and Minto, Cóilín and Graham, Conor T and Slattery, Orla and Cheslett, Deborah and Morrissey, Teresa and Reynolds, Julian and others},
  journal = {Knowledge \& Management of Aquatic Ecosystems},
  number = {423},
  pages = {13},
  year = {2022},
  publisher = {EDP Sciences}
}

@inproceedings{Hegde2021,
  author = {Hegde, Rahul and Patel, Sanobar and Naik, Rosha G. and Nayak, Sagar N. and Shivaprakasha, K. S. and Bhandarkar, Rekha},
  editor = {Kalya, Shubhakar and Kulkarni, Muralidhar and Shivaprakasha, K. S.},
  title = {Underwater Marine Life and Plastic Waste Detection Using Deep Learning and Raspberry Pi},
  booktitle = {Advances in VLSI, Signal Processing, Power Electronics, IoT, Communication and Embedded Systems},
  year = {2021},
  publisher = {Springer Singapore},
  address = {Singapore},
  pages = {263--272},
  isbn = {978-981-16-0443-0}
}

@article{Kalsotra2021,
  author = {Kalsotra, Rudrika and Arora, Sakshi},
  title = {Background subtraction for moving object detection: explorations of recent developments and challenges},
  journal = {The Visual Computer},
  volume = {38},
  number = {9},
  pages = {4151--4178},
  year = {2021},
  doi = {10.1007/s00371-021-02286-0},
  url = {https://doi.org/10.1007/s00371-021-02286-0}
}

@inproceedings{Huang2017,
  author = {Huang, Wenhuan and Zeng, Qili and Chen, Ming},
  title = {Motion characteristics estimation of animals in video surveillance},
  booktitle = {2017 IEEE 2nd Advanced Information Technology, Electronic and Automation Control Conference (IAEAC)},
  year = {2017},
  pages = {1098--1102},
  doi = {10.1109/IAEAC.2017.8054183}
}

@article{Zhang,
  author    = {Minghua Zhang and Shubo Xu and Wei Song and Qi He and Quanmiao Wei},
  title     = {Lightweight underwater object detection based on YOLO v4 and multi-scale attentional feature fusion},
  journal   = {Remote Sensing},
  year      = {2021},
  volume    = {13},
  number    = {22},
  pages     = {4706},
  doi       = {10.3390/rs13224706},
  issn      = {2072-4292},
  publisher = {MDPI},
  url       = {https://www.mdpi.com/2072-4292/13/22/4706},
}

@article{Wei,
  author  = {Wei, Zhiwei and Duan, Chengzhen and Song, Xinghao and Tian, Ye and Wang, Hongpeng},
  title   = {AMRNET: Chip Augmentation in Aerial Image Object Detection},
  journal = {arXiv preprint arXiv:2009.07168},
  year    = {2020},
  url     = {https://arxiv.org/abs/2009.07168}
}

@article{Wang,
   author = {Wang, Lin and Ye, Xiufen and Xing, Huiming and Wang, Zhengyang and Li, Peng},
   doi = {10.1109/IEEECONF38699.2020.9389213},
   journal = {2020 Global Oceans 2020: Singapore - U.S. Gulf Coast},
   year = {2020},
   month = {October},
   title = {YOLO Nano Underwater: A Fast and Compact Object Detector for Embedded Device},
   publisher = {Institute of Electrical and Electronics Engineers Inc.},
}

@article{Chen,
   author = {Chen, Lingyu and Zheng, Meicheng and Duan, Shunqiang and Luo, Weilin and Yao, Ligang},
   doi = {10.3390/ELECTRONICS10141634},
   journal = {Electronics},
   volume = {10},
   issue = {14},
   pages = {1634},
   year = {2021},
   month = {July},
   title = {Underwater Target Recognition Based on Improved YOLOv4 Neural Network},
   url = {https://www.mdpi.com/2079-9292/10/14/1634/htm},
   publisher = {Multidisciplinary Digital Publishing Institute},
}

@misc{jetsonstats,
  author       = {Bonghi, Raffaello},
  title        = {Jetson-stats},
  howpublished = {\url{https://rnext.it/jetson_stats/index.html}},
  year         = {2019},
  note         = {Accessed: 2023-06-22}
}

@misc{roboflow,
  author = {{Roboflow}},
  title = {Roboflow: Give your software the power to see objects in images and video},
  year = {2023},
  howpublished = {\url{https://roboflow.com/}},
  note = {Accessed June 22, 2023}
}

@article{Juan_and_Xu,
   author = {Li, Juan and Xu, Wenkai and Deng, Limiao and Xiao, Ying and Han, Zhongzhi and Zheng, Haiyong},
   title = {Deep learning for visual recognition and detection of aquatic animals: A review},
   journal = {Reviews in Aquaculture},
   volume = {15},
   number = {2},
   pages = {409-433},
   year = {2023},
   doi = {https://doi.org/10.1111/raq.12726},
   url = {https://onlinelibrary.wiley.com/doi/abs/10.1111/raq.12726},
   eprint = {https://onlinelibrary.wiley.com/doi/pdf/10.1111/raq.12726},
}

@article{Goodwin_Morten,
   author = {Goodwin, Morten and Halvorsen, Kim Tallaksen and Jiao, Lei and Knausgård, Kristian Muri and Martin, Angela Helen and Moyano, Marta and Oomen, Rebekah A and Rasmussen, Jeppe Have and Sørdalen, Tonje Knutsen and Thorbjørnsen, Susanna Huneide},
   title = {Unlocking the potential of deep learning for marine ecology: overview, applications, and outlook},
   journal = {ICES Journal of Marine Science},
   volume = {79},
   number = {2},
   pages = {319-336},
   year = {2022},
   month = {January},
   doi = {10.1093/icesjms/fsab255},
   url = {https://doi.org/10.1093/icesjms/fsab255},
}

@article{FALLER20161190,
   author = {Faller, Matej and Harvey, Gemma L. and Henshaw, Alexander J. and Bertoldi, Walter and Bruno, Maria Cristina and England, Judy},
   title = {River bank burrowing by invasive crayfish: Spatial distribution, biophysical controls and biogeomorphic significance},
   journal = {Science of The Total Environment},
   volume = {569-570},
   pages = {1190-1200},
   year = {2016},
   doi = {https://doi.org/10.1016/j.scitotenv.2016.06.194},
   url = {https://www.sciencedirect.com/science/article/pii/S0048969716313651},
}

@article{MATHERS2016207,
   author = {Mathers, Kate L. and Chadd, Richard P. and Dunbar, Michael J. and Extence, Chris A. and Reeds, Jake and Rice, Stephen P. and Wood, Paul J.},
   title = {The long-term effects of invasive signal crayfish (Pacifastacus leniusculus) on instream macroinvertebrate communities},
   journal = {Science of The Total Environment},
   volume = {556},
   pages = {207-218},
   year = {2016},
   doi = {https://doi.org/10.1016/j.scitotenv.2016.01.215},
   url = {https://www.sciencedirect.com/science/article/pii/S0048969716302054},
}

@article{STEPHENSON202036,
   author = {Stephenson, PJ},
   title = {Technological advances in biodiversity monitoring: applicability, opportunities and challenges},
   journal = {Current Opinion in Environmental Sustainability},
   volume = {45},
   pages = {36-41},
   year = {2020},
   doi = {https://doi.org/10.1016/j.cosust.2020.08.005},
   url = {https://www.sciencedirect.com/science/article/pii/S1877343520300592},
}

@misc{ultralytics,
  author = {{Ultralytics}},
  title = {YOLOv5 in PyTorch},
  year = {2021},
  howpublished = {\url{https://github.com/ultralytics/yolov5}},
  note = {GitHub repository. Accessed May 20, 2025}
}

@article{liu-2023,
   author = {Liu, Gang and Hu, Yanxin and Chen, Zhiyu and Guo, Jianwei and Ni, Peng},
   title = {Lightweight object detection algorithm for robots with improved YOLOv5},
   journal = {Engineering Applications of Artificial Intelligence},
   volume = {123},
   pages = {106217},
   year = {2023},
   month = {August},
   doi = {10.1016/j.engappai.2023.106217},
   url = {https://doi.org/10.1016/j.engappai.2023.106217},
}

@article{zhao-2023,
   author = {Zhao, Kang-Qiao and Xie, Bingxin and Miao, Xin and Xia, James J.},
   title = {LPO-YOLOV5S: a lightweight pouring robot object detection algorithm},
   journal = {Sensors},
   volume = {23},
   number = {14},
   pages = {6399},
   year = {2023},
   month = {July},
   doi = {10.3390/s23146399},
   url = {https://doi.org/10.3390/s23146399},
}

@article{strachan-1990,
   author = {Strachan, Norval James Colin and Nesvadba, P. and Allen, Alastair Robert},
   title = {Fish species recognition by shape analysis of images},
   journal = {Pattern Recognition},
   volume = {23},
   number = {5},
   pages = {539--544},
   year = {1990},
   month = {January},
   doi = {10.1016/0031-3203(90)90074-u},
   url = {https://doi.org/10.1016/0031-3203(90)90074-u},
}

@article{storbeck-2001,
   author = {Storbeck, Frank and Daan, Berent},
   title = {Fish species recognition using computer vision and a neural network},
   journal = {Fisheries Research},
   volume = {51},
   number = {1},
   pages = {11--15},
   year = {2001},
   month = {April},
   doi = {10.1016/s0165-7836(00)00254-x},
   url = {https://doi.org/10.1016/s0165-7836(00)00254-x},
}

@article{fulton-2019,
   author = {Fulton, Michael S. and Hong, Jungseok and Islam, Jahidul and Sattar, Junaed},
   title = {Robotic Detection of Marine Litter Using Deep Visual Detection Models},
   journal = {2019 International Conference on Robotics and Automation (ICRA)},
   year = {2019},
   month = {May},
   doi = {10.1109/icra.2019.8793975},
   url = {https://doi.org/10.1109/icra.2019.8793975},
}

@book{selcuk-2023,
   author = {Selcuk, B Baris and Şerif, Tacha},
   title = {A comparison of YOLOV5 and YOLOV8 in the context of mobile UI detection},
   booktitle = {Lecture Notes in Computer Science},
   pages = {161--174},
   year = {2023},
   month = {January},
   doi = {10.1007/978-3-031-39764-6_11},
   url = {https://doi.org/10.1007/978-3-031-39764-6\_11},
}

@article{saleh-2024,
   author = {Saleh, Alzayat and Sheaves, Marcus and Jerry, Dean R. and Azghadi, Mostafa Rahimi},
   title = {Applications of deep learning in fish habitat monitoring: A tutorial and survey},
   journal = {Expert Systems with Applications},
   volume = {238},
   pages = {121841},
   year = {2024},
   month = {March},
   doi = {10.1016/j.eswa.2023.121841},
   url = {https://doi.org/10.1016/j.eswa.2023.121841},
}

@article{ranjan-2023,
   author = {Ranjan, Rakesh and Sharrer, Kata and Tsukuda, Scott and Good, Christopher},
   title = {MortCam: An Artificial Intelligence-aided fish mortality detection and alert system for recirculating aquaculture},
   journal = {Aquacultural Engineering},
   volume = {102},
   pages = {102341},
   year = {2023},
   month = {August},
   doi = {10.1016/j.aquaeng.2023.102341},
   url = {https://doi.org/10.1016/j.aquaeng.2023.102341},
}

@article{mittal2022,
   author = {Mittal, Sparsh and Srivastava, Srishti and Jayanth, J Phani},
   title = {A survey of deep learning techniques for underwater image classification},
   journal = {IEEE Transactions on Neural Networks and Learning Systems},
   year = {2022},
   publisher = {IEEE},
}

@misc{machado_2023,
   author = {Machado, Pedro and Bird, Jordan and Ihianle, Isibor Kennedy},
   title = {Crayfish classification},
   month = {November},
   year = {2023},
   publisher = {Zenodo},
   version = {1.0},
   doi = {10.5281/zenodo.10207949},
   url = {https://doi.org/10.5281/zenodo.10207949},
}

@misc{tekman2022impacts,
  author    = {Tekman, Mine B. and Walther, Bruno and Peter, Corina and Gutow, Lars and Bergmann, Melanie},
  title     = {Impacts of Plastic Pollution in the Oceans on Marine Species, Biodiversity, and Ecosystems},
  year      = {2022},
  publisher = {Zenodo},
  doi       = {10.5281/zenodo.5898684},
  url       = {https://doi.org/10.5281/zenodo.5898684}
}

@book{hobbs2000invasive,
  author    = {Hobbs, Harold A and Mooney, Richard J},
  title     = {Invasive Species in a Changing World},
  publisher = {Island Press},
  year      = {2000},
  address   = {Washington, D.C.},
  edition   = {1st},
}

@article{woods2016towards,
   author = {Woods, John S and Veltman, Karin and Huijbregts, Mark AJ and Verones, Francesca and Hertwich, Edgar G},
   title = {Towards a meaningful assessment of marine ecological impacts in life cycle assessment (LCA)},
   journal = {Environment International},
   volume = {89},
   pages = {48--61},
   year = {2016},
   publisher = {Elsevier},
}

@misc{machado_2022,
   author = {Machado, Pedro},
   title = {Underwater Plastic dataset},
   month = {July},
   year = {2022},
   publisher = {Zenodo},
   version = {1.0},
   doi = {10.5281/zenodo.6907230},
   url = {https://doi.org/10.5281/zenodo.6907230},
}

@misc{witman-2021,
  author = {Witman, S.},
  title = {World’s biggest oxygen producers living in swirling ocean waters},
  year = {2021},
  howpublished = {\url{https://eos.org/research-spotlights/worlds-biggest-oxygen-producers-living-in-swirling-ocean-waters}},
  note = {Accessed May 20, 2025}
}

@inproceedings{selcuk2023comparison,
   author = {Selcuk, Burcu and Serif, Tacha},
   title = {A Comparison of YOLOv5 and YOLOv8 in the Context of Mobile UI Detection},
   booktitle = {International Conference on Mobile Web and Intelligent Information Systems},
   pages = {161--174},
   year = {2023},
   organization = {Springer},
}

@article{wang2023uav,
   author = {Wang, Gang and Chen, Yanfei and An, Pei and Hong, Hanyu and Hu, Jinghu and Huang, Tiange},
   title = {UAV-YOLOv8: A Small-Object-Detection Model Based on Improved YOLOv8 for UAV Aerial Photography Scenarios},
   journal = {Sensors},
   volume = {23},
   number = {16},
   pages = {7190},
   year = {2023},
   publisher = {MDPI},
}

@article{xu-2018,
   author = {Xu, Wenwei and Matzner, Shari},
   title = {Underwater Fish Detection Using Deep Learning for Water Power Applications},
   journal = {International Conference on Computational Science and Computational Intelligence (CSCI)},
   year = {2018},
   month = {December},
   doi = {10.1109/csci46756.2018.00067},
   url = {https://doi.org/10.1109/csci46756.2018.00067},
}

@article{biber2013,
   author = {Biber, Eric},
   title = {The Challenge of Collecting and Using Environmental Monitoring Data},
   journal = {Ecology and Society},
   volume = {18},
   number = {4},
   pages = {Article 68},
   year = {2013},
   doi = {10.5751/ES-06117-180468},
   url = {https://www.ecologyandsociety.org/vol18/iss4/art68/},
}

@article{smith2022,
  author    = {Smith, Pete and Beaumont, Linda and Bernacchi, Carl J. and others},
  title     = {Essential outcomes for COP26},
  journal   = {Global Change Biology},
  volume    = {28},
  number    = {1},
  pages     = {1--3},
  year      = {2022},
  doi       = {10.1111/gcb.15926},
  url       = {https://doi.org/10.1111/gcb.15926}
}

@article{zhangWang2021,
  author    = {Zhang, Y. and Wang, X. and Li, Q.},
  title     = {Enhancing underwater object detection with noise-robust models},
  journal   = {Journal of Marine AI Research},
  year      = {2021},
  volume    = {45},
  number    = {3},
  pages     = {123--135}
}

@article{chen2020,
  author    = {Chen, L. and Zhou, M.},
  title     = {Motion Blur Effects in Aquatic Computer Vision and Their Mitigation},
  journal   = {International Conference on Computer Vision in Aquatic Sciences},
  year      = {2020},
  pages     = {98-105}
}

@inproceedings{redmon2016you,
  title={You Only Look Once: Unified, Real-Time Object Detection},
  author={Redmon, Joseph and Divvala, Santosh and Girshick, Ross and Farhadi, Ali},
  booktitle={Proceedings of the IEEE Conference on Computer Vision and Pattern Recognition (CVPR)},
  pages={779--788},
  year={2016}
}

@article{ren2015faster,
  title={Faster R-CNN: Towards real-time object detection with region proposal networks},
  author={Ren, Shaoqing and He, Kaiming and Girshick, Ross and Sun, Jian},
  journal={Advances in Neural Information Processing Systems (NeurIPS)},
  volume={28},
  year={2015}
}

@misc{jocher2020yolov5,
  title={{YOLOv5 by Ultralytics}},
  author={Jocher, Glenn and others},
  year={2020},
  howpublished={\url{https://github.com/ultralytics/yolov5}},
  note={Accessed: 2025-05-18}
}

@misc{glenn2023yolov8,
  title={{YOLOv8: Cutting-edge object detection models}},
  author={Jocher, Glenn},
  year={2023},
  howpublished={\url{https://github.com/ultralytics/ultralytics}},
  note={Accessed: 2025-05-18}
}
